\documentclass[conference]{IEEEtran}
\IEEEoverridecommandlockouts
\usepackage{cite}
\usepackage{amsmath,amssymb,amsfonts}
\usepackage{algorithmic}
\usepackage{graphicx}
\usepackage{textcomp}
\usepackage{xcolor}
\usepackage{booktabs}
\usepackage{multirow}
\usepackage{threeparttable}
\usepackage{comment}
\usepackage{cite}

\usepackage[whole]{bxcjkjatype}

\newcommand{\tabref}[1]{Table\hspace{1mm}\ref{#1}}

\def\BibTeX{{\rm B\kern-.05em{\sc i\kern-.025em b}\kern-.08em
    T\kern-.1667em\lower.7ex\hbox{E}\kern-.125emX}}

\makeatletter

\def\ps@IEEEtitlepagestyle{%
  \def\@oddfoot{\mycopyrightnotice}%
  \def\@evenfoot{}%
}
\def\mycopyrightnotice{%
  {\footnotesize 978-1-7281-0858-2/19/\$31.00 © 2019 IEEE\hfill}
  \gdef\mycopyrightnotice{}
}

\begin{document}

\title{AOP: An Anti-overfitting Pretreatment \\
for Practical Image-based Plant Diagnosis
}

\author{
\IEEEauthorblockN{
    Takumi Saikawa\IEEEauthorrefmark{1},
    Quan Huu Cap\IEEEauthorrefmark{1},
    Satoshi Kagiwada\IEEEauthorrefmark{2},
    Hiroyuki Uga\IEEEauthorrefmark{3},
    Hitoshi Iyatomi\IEEEauthorrefmark{1}
    }
\IEEEauthorblockA{
Email: takumi.saikawa.5x@stu.hosei.ac.jp huu.quan.cap.78@stu.hosei.ac.jp \\ kagiwada@hosei.ac.jp uga.hiroyuki@pref.saitama.lg.jp iyatomi@hosei.ac.jp \\
\IEEEauthorrefmark{1}Applied Informatics, Graduate School of Science and Engineering, Hosei University, Tokyo, Japan}
\IEEEauthorblockA{
\IEEEauthorrefmark{2}Clinical Plant Science, Faculty of Bioscience and Applied Chemistry, Hosei University, Tokyo, Japan}
\IEEEauthorblockA{
\IEEEauthorrefmark{3}Saitama Agricultural Technology Research Center, Saitama, Japan}
}
\maketitle


\begin{abstract}
In image-based plant diagnosis, clues related to diagnosis are often unclear, and the other factors such as image backgrounds often have a significant impact on the final decision.
As a result, overfitting due to latent similarities in the dataset often occurs, and the diagnostic performance on real unseen data (e,g. images from other farms) is usually dropped significantly.
However, this problem has not been sufficiently explored, since many systems have shown excellent diagnostic performance due to the bias caused by the similarities in the dataset.
In this study, we investigate this problem with experiments using more than 50,000 images of cucumber leaves, and propose an anti-overfitting pretreatment (AOP) for realizing practical image-based plant diagnosis systems.
The AOP detects the area of interest (leaf, fruit etc.) and performs brightness calibration as a preprocessing step.
The experimental results demonstrate that our AOP can improve the accuracy of diagnosis for unknown test images from different farms by 12.2\% in a practical setting.

\end{abstract}

\begin{IEEEkeywords}
automated plant diagnosis, preprocessing, segmentation
\end{IEEEkeywords}

\begin{table*}[t]
\caption{DISEASE DATASET USED IN THIS STUDY}
\centering
\label{tab:dataset}
\begin{tabular}{@{}clrrr@{}}
\toprule
\multicolumn{2}{c}{\multirow{2}{*}{Diseases}}     & \multicolumn{3}{c}{Disease classification dataset}                                                     \\ \cmidrule(l){3-5}
\multicolumn{2}{c}{}                              & \multicolumn{1}{c}{Training} & \multicolumn{1}{l}{Validation} & \multicolumn{1}{c}{Test (Different farm)} \\ \cmidrule(r){1-2} \cmidrule(lr){3-3} \cmidrule(lr){4-4} \cmidrule(lr){5-5}
\multirow{4}{*}{Viral diseases}  & MYSV           & 7,448                     & 827                            & 2,596                                     \\
                                 & ZYMV           & 9,189                     & 1,021                          & 3,364                                     \\
                                 & CMV            & 4,109                     & 456                            & 563                                       \\
                                 & WMV            & 2,511                     & 279                            & 306                                       \\ \cmidrule(r){1-2} \cmidrule(lr){3-3} \cmidrule(lr){4-4} \cmidrule(lr){5-5}
\multirow{3}{*}{Fungal diseases} & Brown Spot     & 2,014                     & 224                            & 654                                       \\
                                 & Downy Mildew    & 1,311                     & 146                            & 380                                       \\
                                 & Powdery Mildew & 2,204                     & 245                            & 114                                       \\ \cmidrule(r){1-2} \cmidrule(lr){3-3} \cmidrule(lr){4-4} \cmidrule(lr){5-5}
\multicolumn{2}{c}{Healthy}                       & 6,908                     & 768                            & 1,138                                     \\ \cmidrule(r){1-2} \cmidrule(lr){3-3} \cmidrule(lr){4-4} \cmidrule(lr){5-5}
\multicolumn{2}{c}{Total}                         & 35,694                    & 3,966                          & 9,115                                     \\ \bottomrule
\end{tabular}
\end{table*}

\section{Introduction}

Plant health is important to ensure a balance between global food demand and agricultural productivity.
Protecting plants against disease plays a crucial role in the agricultural domain, helping to improve crop yields and increase food security.
In general, the detection of plant diseases is time-consuming, as it requires a great deal of work related to evaluation and treatment by plant pathologists, especially on large farms.
The detection of plant disease via automatic techniques is therefore highly desirable.

Plant leaves are the most common part used to detect disease.
The automated detection of plant diseases from leaf images has been widely studied in the scientific community, and many proposals have achieved promising results using powerful approaches based on deep convolutional neural networks (CNNs) \cite{krizhevsky2012imagenet} and related models.
As a pioneer study, Kawasaki et al. \cite{kawasaki2015basic} developed a three-layer CNN to classify three classes of cucumber disease (two diseases and healthy specimens) using leaf images from a real farm.
Their model achieved an average accuracy of 94.9\% for a wide variety of photographic conditions and complex backgrounds.
Fujita et al. \cite{fujita2016basic} extended their study by including early-stage and ill-conditioned images in order to consider more practical situations.
They achieved an average classification performance of 82.3\% for an eight-class diagnosis scheme for cucumbers.
Wang et al. \cite{wang2017automatic} classified black rot apple diseases using the PlantVillage open dataset \cite{hughes2015open} and obtained an average accuracy of 90.4\%.
Diagnosis studies on tomatoes \cite{durmucs2017disease,atabay2017deep} and several other plant species \cite{sladojevic2016deep} have also demonstrated good classification performance.

More practical diagnosis systems using a wider range of images, including several types of object, have been also proposed. Fuentes et al. \cite{fuentes2017robust} used Faster R-CNN \cite{ren2015faster} and SSD \cite{liu2016ssd} in an object detection and recognition strategy that was state-of-the-art at the time.
They investigated a total of 5,000 annotated tomato field leaf images and achieved a maximum mean average precision of 83.6\%.
Lu et al. \cite{lu2017field} designed a similar framework using a fully convolutional network, and achieved a mean recognition accuracy of 98.0\% on a wheat disease dataset; they claim that their system can also be deployed in mobile applications.

Although the above-mentioned studies have shown good results for disease diagnosis thanks to breakthroughs in deep learning models, one of the remaining issues is the overfitting problem, which significantly reduces the test diagnosis performance, particularly when the test images are collected from a different environment than the training images. Although this is a very serious problem, only a limited number of studies have highlighted this aspect.

Mohanty et al. \cite{mohanty2016using} used a total of 54,306 plant leaf images consisting of 38 classes of crop-disease pairs (14 crop species) from the PlantVillage dataset, and achieved a classification accuracy of 99.3\%.
However, they also noted that the accuracy dropped to around 31\% when diagnosis was carried out with different test settings. In \cite{fuentes2017robust}, a pre-trained CNN attained a success rate of 99.5\% when identifying 58 classes (plant types and diseases) based on a dataset drawn from both the laboratory and cultivated fields.
However, this rate dropped to about 33\% when the model was trained solely with laboratory-condition images and tested with cultivation-condition images.
This problem also arises when the diagnostic system is trained and tested on different in-farm leaf datasets.
Cap et al. developed a practical plant diagnosis system that is capable of using practical wide-angle images taken by surveillance cameras \cite{cap2018end}.
In the images used, several dozen leaves were present, and were heavily overlapped.
Their diagnosis model was composed of a VGG-based network fine-tuned with more than 30,000 cucumber leaf images and achieved a mean accuracy of 97.4\% on a test dataset from the same farm; however, this dropped to 65.8\% when they tested it on images from a different farm.

In plant diagnosis tasks, the appearance of symptoms is not always clear, and thus the influence of other factors such as variance in the background and image condition is much larger than in general object recognition tasks.
Thus, overfitting becomes important.
Fujita et al. \cite{fujita2018practical} visualized the key regions of diagnostic evidence for their cucumber diagnosis system using GradCAM \cite{selvaraju2017grad}. Although their model achieved an average accuracy of 93.6\%, they claimed that it sometimes responded to the background areas instead of the target leaf regions, due to the overfitting problem.

These results indicate that some sort of standardization process for images is required.
In particular, when plants are grown in a controlled environment to ensure the quality of the training labels, the background diversity tends to be limited.
For instance, a set of images infected with a certain type of disease may all have a specific background.
For this reason, we believe that the introduction of an efficient preprocessing step to segment out only the region of interest (RoI) is needed to reduce the abovementioned overfitting problem, which is particularly prominent in plant diagnosis.

Segmentation itself has a long research history, and the robust extraction of a RoI against complex backgrounds is generally difficult.
Recently, several robust methods based on deep learning have been proposed and have been shown to have outstanding capabilities. For instance, U-net \cite{ronneberger2015u} and its improved version called pix2pix \cite{isola2017image} do not require any knowledge of the target domain, and have obtained very reasonable segmentation performance just by learning pairs of training images and their segmentation areas. In the past, segmentation of an RoI as part of preprocessing was an essential step in most image recognition or computer vision tasks; however, in the era of deep learning, the need for this step has decreased, and it is generally omitted except in some medical tasks. For the reasons described above, a sophisticated region extraction process is now required in practical image-based plant diagnosis tasks.

In this study, in order to address the serious overfitting problem, we propose an anti-overfitting pretreatment (AOP) to detect the area of interest (i.e. leaf, fruit, etc.) as a preprocessing step for image-based plant diagnosis systems.

To the best of our knowledge, this is the first study to conduct segmentation of the area of interest for a plant diagnosis task and to confirm the validity using real farm images.

\section{Materials and methods}

\begin{figure}[t]
  \begin{center}
    \includegraphics[width=\linewidth]{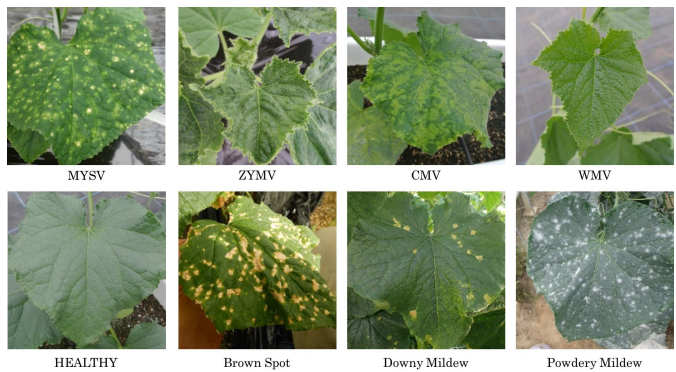}
    \caption{
        Typical examples of seven diseases and a healthy leaf.
        Each image is labeled as one of four types of viral disease (MYSV, ZYMV, CMV, and WMV), three types of fungal disease (Brown Spot, Downy Mildew, and Powdery Mildew) or a healthy leaf.
    }
    \label{fig:sample}
  \end{center}
\end{figure}

\subsection{Dataset}
In this study, we use two types of cucumber leaf datasets.
The first is the segmentation dataset, which consists of a total of 8,000 cucumber leaf images and their corresponding leaf region pixel-level annotated ground-truths.
This dataset is used to train the core network of our proposed AOP.
Note that we collected only images in good condition for our segmentation dataset (i.e. we removed images that were too bright or too dark, and other ill-conditioned images, in advance).
Because this data set consists of images with various aspect ratios, we apply a square center crop with the short side of the image as the side length.
We then resize them to $316 \times 316$ pixels.

The second is a classification dataset consisting of 48,775 single-leaf cucumber images, as summarized in \tabref{tab:dataset}.
These images were labeled by experts from agricultural technology research centers in six prefectures of Japan, and each image contains a single cucumber leaf in the middle.
Each image in this dataset is of one of four types of viral disease (MYSV, ZYMV, CMV, and WMV)\footnote{MYSV: Melon yellow spot virus, ZYMV: Zucchini yellow mosaic virus, CMV: Cucumber mosaic virus, WMV: Watermelon mosaic virus}, three fungal diseases (Brown Spot, Downy Mildew, and Powdery Mildew) or a healthy leaf.
This classification dataset has three subsets, i.e. the $\text{disease}_\text{training}$, the $\text{disease}_\text{validation}$, and the $\text{disease}_\text{test}$, which are used to train/validate and test the final disease diagnosis networks (details of this are given later).
The $\text{disease}_\text{training}$ set contains 35,694 images, the $\text{disease}_\text{validation}$ set 3,966 images, and the $\text{disease}_\text{test}$ set 9,115 images.
All the images are resized to $224 \times 224$ pixels.
Note that the training/validation sets were taken from the same farm, but that the testing set was taken from completely different farms.
Fig. \ref{fig:sample} shows typical example images for each category.

\begin{figure}[t]
  \begin{center}
    \includegraphics[width=8cm]{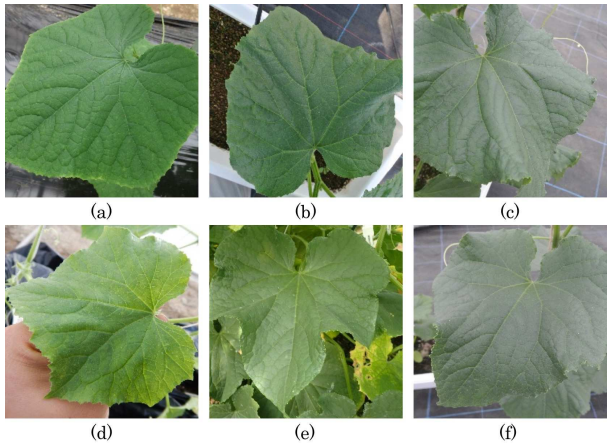}
    \caption{
        Early symptoms of (a) MYSV; (b) CMV; (c) WMV; (d) Brown Spot; (e) Powdery Mildew; and (f) a healthy leaf.
        Early symptoms are almost indistinguishable from healthy leaves and the early stages of other diseases.
    }
    \label{fig:early_symptons}
  \end{center}
\end{figure}

It should also be mentioned that the condition of some of the images in this classification dataset is not good compared to the segmentation dataset.
Moreover, this dataset contains many images of early-stage disease, i.e. MYSV, CMV, WMV, Brown Spot and Powdery Mildew.
These early-stage images are almost indistinguishable, both from each other and from healthy images.
Powdery Mildew is the class containing most of the early-stage images (early-stage images account for more than 25\% of the total images for Powdery Mildew disease).
Fig. \ref{fig:early_symptons} also shows examples of these early cases.

\subsection{The proposed anti-overfitting pretreatment (AOP)}
The key features of our proposed anti-overfitting pretreatment (AOP) are a robust segmentation process of the RoI (i.e. the leaf shape) and image brightness calibration.
These functions reduce the serious negative effects of the background in terms of causing the overfitting problem.
In our experiment, we examine the effect of the AOP on diagnosis of cucumber disease based on leaf pictures taken from real farms.

\begin{figure}[t]
  \begin{center}
    \includegraphics[width=\linewidth]{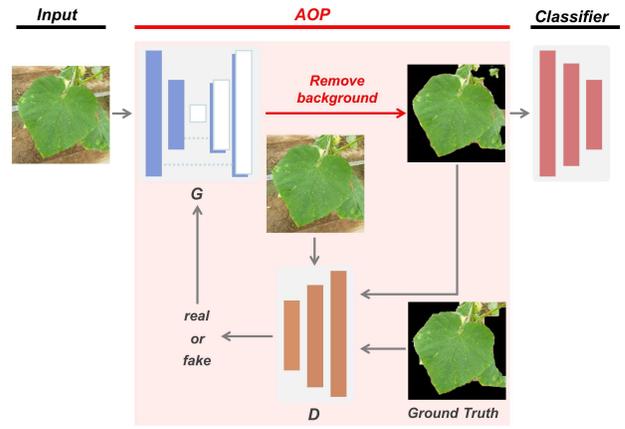}
    \caption{
        AOP network architecture.
        Our AOP network is based on the pix2pix architecture, and is composed of a generator and a discriminator.
    }
    \label{fig:structure}
  \end{center}
\end{figure}

Our AOP network is based on the pix2pix architecture \cite{isola2017image}, and is composed of a generator and a discriminator.
Fig. \ref{fig:structure} shows our plant diagnosis schema with the proposed AOP.
The generator is based on the U-net architecture \cite{ronneberger2015u}, and the encoder and decoder are basically the same as in the original pix2pix model.
Specifically, the encoder consists of eight convolution and eight deconvolution layers, while the decoder has five convolutional layers.
Since the vanilla U-net often generates checkerboard artifacts \cite{odena2016deconvolution} arising from the deconvolution process in the image generator, we added an unpooling layer after each convolution layer to avoid this.

Pix2pix has been shown to have excellent segmentation and image translation capabilities.
Since pix2pix is used for image translation, we believe that it can be used to improve the condition of bad images, such as those with poor lighting conditions.
Here, our generator carries out image segmentation and also improves the image condition.
Our images are taken under different photographic conditions on different farms, meaning that poorly conditioned images are included.
This can lead to incorrect segmentation and lower disease diagnostic performance, since important parts of some images are missing.
In order to deal with this problem, we apply brightness calibration to the input images using gamma correction to train our AOP models.
We assume that this brightness calibration allows our model to generate segmented images with better photographic conditions.

To achieve this, we apply gamma correction after performing several online data augmentation techniques to train the AOP, i.e. image rotation with a step size of 90°, image mirroring, and random cropping to a size of $256 \times 256$ pixels.
Gamma correction is used, with gamma values randomly selected from [0.8, 1.0, 1.5, 2.0, 2.2].
In addition to the mean absolute error (MAE) content loss used in pix2pix, we also introduce structural similarity (SSIM) \cite{wang2004image} to our models, since improving the target structure is expected to improve the diagnostic performance of the final disease diagnosis system.
In our experiment, we compare the difference in diagnosis performance through several AOP models.

\subsection{Disease classification networks}
The final aim of this study is to prove the effectiveness of our proposed AOP against the overfitting problem when testing the diagnostic systems on unknown datasets.
We therefore design different disease classification networks and test them on the unknown $\text{disease}_\text{test}$ dataset.
These networks are based on the VGG-16 network \cite{simonyan2014very} and are pre-trained with the ImageNet dataset \cite{deng2009imagenet}; we fine-tune them on the $\text{disease}_\text{training}$ dataset.

We use small numbers of neurons (1024 and 32) in the FC layers before the eight-class output in order to avoid model overfitting. The details of the training disease classification networks will be described in the next section.

\subsection{Details of training the networks} 
\subsubsection{Training AOP models}
In the training of the AOP models, the data augmentation techniques described earlier are performed on the fly.
We randomly select 80\% of the segmentation dataset for training and the remaining 20\% is used for testing the AOP models.
In this work, we train three AOP models with different loss functions for comparison purposes: the ${\rm AOP}_{\rm MAE_{\rm prob}}$, which generates a segmentation mask with the original MAE loss from pix2pix as its content loss (i.e. the output is in a probabilistic form and the thresholding forms the mask); the ${\rm AOP}_{\rm MAE}$, with a MAE content loss that directly generates pixel values; and the ${\rm AOP}_{\rm SSIM}$, which uses the structural similarity as its loss function.

We apply an Adam optimizer \cite{kingma2014adam} with $\alpha=0.0002$ and $\beta=0.5$ for both the generator and discriminator of the AOP models. The minibatch was set to 16, and the training process finished after 100 epochs.

\subsubsection{Training disease classification networks}
In this experiment, we train four disease classification networks in order to compare the final diagnostic performance.
The first classifier is trained, validated, and tested with the original versions of the $\text{disease}_\text{training}$, $\text{disease}_\text{validation}$, and the $\text{disease}_\text{test}$ datasets, respectively.
The other three classifiers are trained, validated, and tested on the segmented versions of these datasets produced by the three AOP models, as described in the previous subsection.

To train these networks, the online data augmentation techniques described above were applied, but with a rotation angle step size of 10° and randomly selected gamma values of [0.5, 1.0, 1.5].
Our disease classifiers were optimized using the momentum stochastic gradient descent \cite{qian1999momentum} with the minibatch set to 32, and the initial learning rate was 0.001.
The training process was also stopped after 100 epochs.
\section{Experimental results}

\subsection{Segmentation performance of the AOP networks}
\begin{figure}[t]
  \begin{center}
    \includegraphics[width=\linewidth]{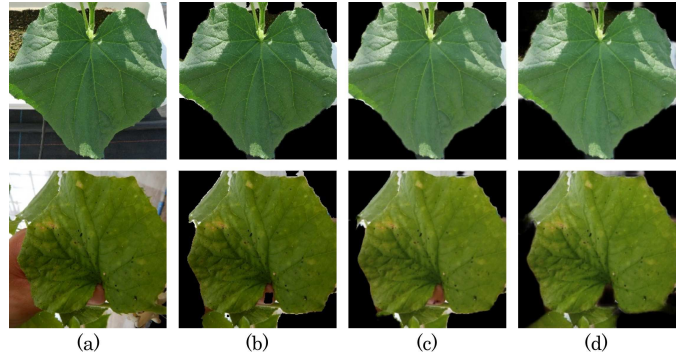}
    \caption{
        Examples of (a) an input image; (b) an image of the extracted leaf area ${\rm AOP}_{\rm MAE_{prob}}$; (c) an image generated by ${\rm AOP}_{\rm MAE}$; and (d) an image generated by ${\rm AOP}_{\rm SSIM}$.
        It can be seen that the leaf region is extracted properly.
    }
    \label{fig:leaves}
  \end{center}
\end{figure}

\tabref{tab:segmentation_result} compares the segmentation performance for the test dataset (20\% of the segmentation dataset), and Fig. \ref{fig:leaves} shows examples of results of segmented leaves.
We use the precision, recall, and F1-score to evaluate the segmentation performance for the three AOP models.
The segmentation results of ${\rm AOP}_{\rm MAE_{prob}}$ and ${\rm AOP}_{\rm MAE}$ show that the leaf region is extracted properly.

Note that we choose a threshold value of 0.5 to generate the mask from the ${\rm AOP}_{\rm MAE_{prob}}$ model.
All of the AOP models achieve high segmentation performance, and the best F1-score of 98.1\% is obtained from the proposed ${\rm AOP}_{\rm SSIM}$.
The results confirm that these segmentation models are strong enough for background removal and are expected to improve the disease diagnostic performance of the subsequent diagnosis systems.

\begin{table}[t]
\centering
\caption{Segmentation performance with AOP networks}
\label{tab:segmentation_result}
\begin{tabular}{@{}lccc@{}}
\toprule
                             & Precision [\%] & Recall [\%] & F1-score [\%] \\ \cmidrule(lr){1-1} \cmidrule(lr){2-2} \cmidrule(lr){3-3} \cmidrule(lr){4-4}
${\rm AOP}_{\rm MAE_{prob}}$ & 100.0          & 95.8        & 95.8          \\
${\rm AOP}_{\rm MAE}$        & 98.4           & 97.1        & 97.8          \\
\textbf{(Proposed)} ${\rm AOP}_{\rm SSIM}$    & 98.6           & 97.5        & \textbf{98.1}          \\ \bottomrule
\end{tabular}
\end{table}

\subsection{Disease classification performance}
The disease classification performance was evaluated and compared, both with and without the AOP models.
\tabref{tab:classification_result} summarizes the performance in terms of average accuracy.
While there are only small differences in the classification accuracy between the $\text{disease}_\text{training}$ and $\text{disease}_\text{validation}$ sets, there are significant undesirable differences between the $\text{disease}_\text{validation}$ and $\text{disease}_\text{test}$ sets.
The biggest drop in performance is shown by the classification model without AOP.
The accuracy of this model drops maximally from 97.5\% on $\text{disease}_\text{validation}$ to only 40.3\% on $\text{disease}_\text{test}$ dataset.
In contrast, the classification network with our ${\rm AOP}_{\rm SSIM}$ significantly improves the accuracy on the unknown $\text{disease}_\text{test}$ dataset by 12.2\%, reaching the best performance of 52.5\% classification accuracy.

\begin{figure}[t]
    \centering
    \includegraphics[width=0.8\linewidth]{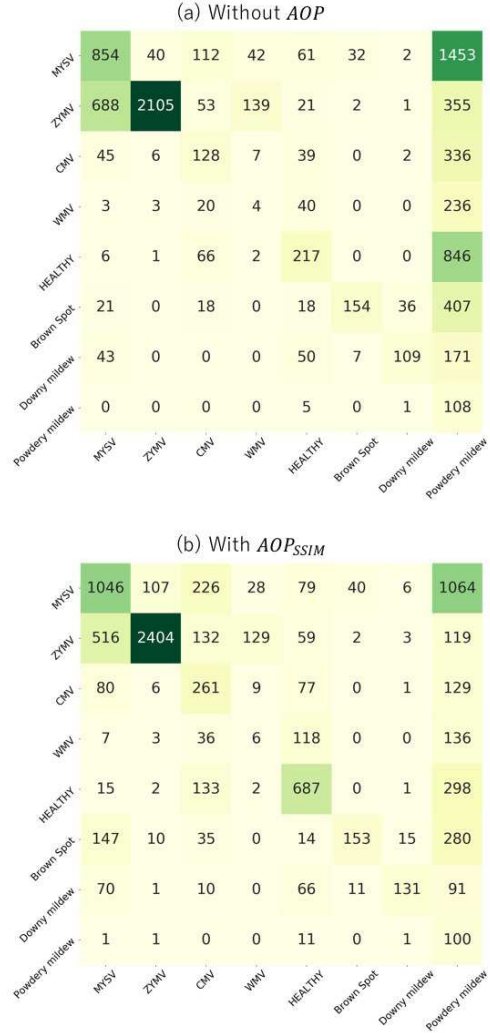}
    \caption{
        Confusion matrices for disease classification networks with and without AOP.
        Although there are a significant number of misclassifications for Powdery Mildew in both cases, the ${\rm AOP}_{\rm SSIM}$ markedly improves the diagnostic performance.
    }
    \label{fig:matrix}
\end{figure}

\begin{figure}[t]
    \centering
    \includegraphics[width=8cm]{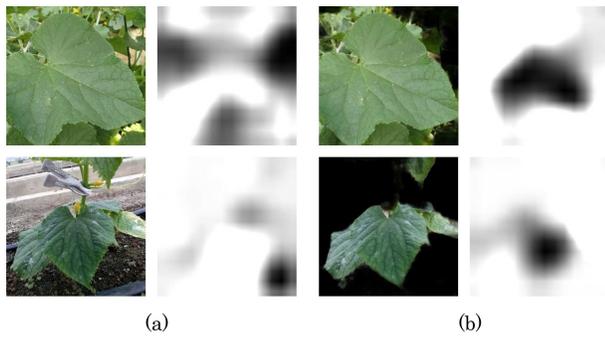}
    \caption{
        Comparison of estimated evidence of diagnosis (a) without AOP; and (b) with the proposed ${\rm AOP}_{\rm SSIM}$.
        In the results with AOP, the highlighted areas overlap the leaf area, while without AOP they do not, as reported in the literature.
    }
    \label{fig:gradcam}
\end{figure}

\begin{table}[t]
\centering
\begin{threeparttable}
\caption{Diagnosis performance with and without AOP}
\label{tab:classification_result}
\begin{tabular}{@{}lccc@{}}
\toprule
\multirow{2}{*}{}                          & \multicolumn{3}{c}{Average Accuracy [\%] (8 Classes)} \\ \cmidrule(l){2-4} 
                                           & $\text{Disease}_\text{training}$   & $\text{Disease}_\text{validation}$  & $\text{Disease}_\text{test}$ \\ \cmidrule(r){1-1} \cmidrule(lr){2-2} \cmidrule(lr){3-3} \cmidrule(l){4-4}
w/o AOP                                    & 98.2            & 97.5                & 40.3          \\
${\rm AOP}_{\rm MAE_{prob}}$               & 98.2            & 97.2                & 48.8          \\
${\rm AOP}_{\rm MAE}$                      & 98.0            & 97.1                & 49.7          \\
\textbf{(Proposed) ${\rm AOP}_{\rm SSIM}$} & 98.0            & 97.4                & \textbf{52.5} \\ \bottomrule
\end{tabular}
\end{threeparttable}
\end{table}
Fig. \ref{fig:matrix} compares the confusion matrices for the disease classification networks (a) without AOP and (b) with ${\rm AOP}_{\rm SSIM}$.
Although there are several misclassifications for Powdery Mildew in both cases, we can confirm that the ${\rm AOP}_{\rm SSIM}$ markedly improves the diagnostic performance.
In addition, we also find that our ${\rm AOP}_{\rm SSIM}$ helps the classification network to focus on the leaf area instead of on its background.
Fig. \ref{fig:gradcam} shows the diagnostic evidence heatmap of the classification network (a) without AOP and (b) with the proposed ${\rm AOP}_{\rm SSIM}$, using the Grad-CAM \cite{selvaraju2017grad} technique.
The darker the heatmap, the stronger the classifier considers that areas.
The results with ${\rm AOP}_{\rm SSIM}$ show that the classification network responds to the leaf area, while the model without AOP responds to the background areas.

\section{Discussion}

These experiments indicate that the diagnosis performance for the test dataset is much lower than for the training/validation dataset in all cases (see \tabref{tab:classification_result}).
The primary reason for this low performance is the misdiagnosis for Powdery Mildew, as shown in the confusion matrix in Fig. \ref{fig:matrix}. We can see that there are numerous false positives for Powdery Mildew compared to other diseases and healthy data.
WMV is also mostly misclassified as healthy or Powdery Mildew.

Since there are 2,204 images of Powdery Mildew in the training set, but almost 25\% of these show early symptoms, our model learns to mistakenly classify the early symptoms of the other diseases of WMV, MYSV, CMV, Brown Spot and healthy images as Powdery Mildew, as they are very similar to each other (see Fig. \ref{fig:early_symptons}).
This problem with early symptoms needs to be addressed, and we leave this for future work.

We confirm that the introduction of AOP draws the classifier's attention to the leaf, rather than being confused by the background, as shown in Fig. \ref{fig:gradcam}, and this significantly improves the classification performance for a completely different dataset.
As shown in \tabref{tab:classification_result}, the accuracy of the disease classifier without AOP for the test data was 40.3\%, while the classifier with ${\rm AOP}_{\rm SSIM}$ achieved an accuracy of 52.5\%, thus improving the overall accuracy by 12.2\%.
This result also supports our hypothesis that SSIM, which captures structural features better, is more effective for the final diagnosis in situations where leaf regions have been properly extracted and where a background that causes disturbance has been removed.

We can also see from the results in \tabref{tab:classification_result} that the ${\rm AOP}_{\rm MAE}$, which acts as a brightness calibrator, improves the diagnosis performance by 0.9\% compared to the ${\rm AOP}_{\rm MAE_{\rm prob}}$.
However, this improvement was smaller than expected.
This is because the brightness gamma correction data augmentation in the diagnosis classifier also contributes to leveraging the final diagnostic performance.

In summary, introducing the AOP eliminates the influence of the background and significantly improves the diagnosis performance. Although the overall performance of 52.5\% for the $\text{disease}_\text{test}$ set is not sufficient for practical systems, mainly due to the limited performance in regard to early-stage symptoms, we have confidence that these results reflect a real system accuracy based on tens of thousands of training images, including a large number of early-stage cases. We will continue to investigate other crucial factors to achieve a practical diagnosis performance.

\section{Conclusion}
In this paper, we confirm the problem caused by latent similarities in image-based plant diagnosis studies. 
We propose an AOP that provides appropriate segmentation of the RoI, in order to realize robust image-based plant diagnosis that reduces serious overfitting problems. With the introduction of the AOP, the final diagnosis performance for cucumber leaves improved by 12.2\% on average for an eight-class classification task.
\section*{Acknowledgment}

This research was supported in part by the Ministry of Education, Science, Sports and Culture, Grant-in-Aid for Fundamental Research (C), 17K08033, 2017–2020.

    \bibliographystyle{IEEEtran}
    \bibliography{ref}

\end{document}